\documentclass[10pt,twocolumn,letterpaper]{article}

\usepackage{cvpr}
\usepackage{times}
\usepackage{epsfig}
\usepackage{graphicx}
\usepackage{amsmath}
\usepackage{amssymb}
\usepackage[table,xcdraw]{xcolor}
\usepackage{bbm}
\usepackage{caption}
\usepackage{subcaption}
\usepackage{enumitem}
\usepackage{tabularx}

\usepackage[pagebackref=true,breaklinks=true,letterpaper=true,colorlinks,bookmarks=false]{hyperref}

\cvprfinalcopy 

\def\cvprPaperID{****} 

\begin{document}

\title{Real-Time Flying Object Detection with YOLOv8}

\author{Dillon Reis*, Jacqueline Hong, Jordan Kupec, Ahmad Daoudi\\
Georgia Institute of Technology\\
{\tt\small dreis7@gatech.edu*, jhong356@gatech.edu, jkupec3@gatech.edu, adaoudi3@gatech.edu}
}

\maketitle

\begin{abstract}
This paper presents a generalized model for real-time detection of flying objects that can be used for transfer learning and further research, as well as a refined model that achieves state-of-the-art results for flying object detection. We achieve this by training our first (generalized) model on a data set containing 40 different classes of flying objects, forcing the model to extract abstract feature representations. We then perform transfer learning with these learned parameters on a data set more representative of ``real world” environments (i.e. higher frequency of occlusion, very small spatial sizes, rotations, etc.) to generate our refined model. Object detection of flying objects remains challenging due to large variances of object spatial sizes/aspect ratios, rate of speed, occlusion, and clustered backgrounds. To address some of the presented challenges while simultaneously maximizing performance, we utilize the current state-of-the-art single-shot detector, YOLOv8, in an attempt to find the best trade-off between inference speed and mean average precision (mAP). While YOLOv8 is being regarded as the new state-of-the-art \cite{YOLOv8Website}, an official paper has not been released as of yet. Thus, we provide an in-depth explanation of the new architecture and functionality that YOLOv8 has adapted. Our final generalized model achieves a mAP50 of 79.2\%, mAP50-95 of 68.5\%, and an average inference speed of 50 frames per second (fps) on 1080p videos. Our final refined model maintains this inference speed and achieves an improved mAP50 of 99.1\% and mAP50-95 of 83.5\%.  
\end{abstract}

\section{Introduction}

Numerous recent events have demonstrated the malicious use of drones. Over the past few months, there have been reports of assassination attempts via drones with small explosive payloads \cite{SuicideDrone}, drug deliveries to state prisons \cite{PrisonDrugs}, and surveillance of the United States (U.S.) Border Patrol by smugglers \cite{BorderPatrol} to exploit weaknesses. While research indicates that drone usage is expected to increase exponentially \cite{DroneMarket}, detection technology has yet to provide reliable and accurate results. Drones and mini unmanned aerial vehicles (UAVs) present a stealth capability and can avoid detection by most modern radar systems due to their small electromagnetic signature. They are also small, highly maneuverable, and omit low levels of noise. This, along with the ease of access, provides a natural incentive for drones to remain an integral part of modern warfare and illegal activities. While methods such as radio and acoustic detection have been proposed as solutions, they are currently known to be inaccurate \cite{Drone-Detection-Using-YOLOv5}. This motivates the integration of a visual detector in any such detection system. The U.S. Border Patrol implements real-time object detection from digital towers to monitor people and motor vehicles \cite{BorderDetection}, but is not currently known to implement drone detection, which may explain the recent undetected illegal patrolling. Drone detection in this environment is challenging due to the cluttered desert background and the distance that drones survey from \cite{BorderDigitalTowers}. The farther the drone is from cameras, the more difficult it will be to detect and classify it, as the object will convey less signal in the input space to the model.\\
\indent Our primary objective is to provide a generalized real-time flying object detection model that can be used by others for transfer learning or further research, as well as a refined model that is ready to use ``out of the box'' for implementation ~\cite{OURCODE}. We define a generalized model as one that has good detection and classification performance on a large number of classes at higher resolutions while maintaining a reasonable frame rate (1080p : 30-60 frames per second). Instead of just training our model on drones, we train on a data set containing 40 different flying object categories to force the model to learn more abstract feature representations of flying objects. Then, we transfer learn these weights on a final data set containing more instances of ``real world`` environments (i.e. higher frequency of occlusion, small spatial sizes, rotations, etc.). This in turn will lead to a more refined, ready-to-implement real-time flying object detection model. To maximize our model's performance, we use the latest state-of-the-art single-shot detector, YOLOv8. Currently, single-stage detectors are the de-facto architecture choice for fast inference speeds. This choice comes at the expense of exchanging the higher accuracy you would typically expect from a two-state detector. While YOLOv8 is being regarded as the new state-of-the-art \cite{YOLOv8Website}, an official paper has yet to be released. This motivates our secondary objective, which is to explain the new architecture and functionality that YOLOv8 has adapted.

\section{Materials and Methods}

Real-time object detection remains challenging due to variances in object spatial sizes and aspect ratios, inference speed, and noise. This is especially true for our use case, as flying objects can change location, scale, rotation, and trajectory very quickly. This conveys the necessity for fast inference speed and thorough model evaluation between low-variance classes, object sizes, rotations, backgrounds, and aspect ratios.

Our initial model is trained on a data set \cite{InitialDataset} comprised of 15,064 images of various flying objects with an 80\% train and 20\% validation split. Each image is labeled with the class number of the object and the coordinates of the edges of the associated bounding box. An image may have more than one object and class, sitting at an average of 1.6 annotated objects per image and a total of 24,769 annotations across all images. The median image ratio is 416x416. The images were pre-processed with auto-orientation, but there were no augmentations applied. The data set represents a long-tailed distribution with the drone (25.2\% of objects), bird (25\%), p-airplane (7.9\%), and c-helicopter (6.3\%) classes taking up the majority of the data set (64.4\%), suffering from a class imbalance. Published on Roboflow with an unnamed author, this data set was generated in 2022, having been downloaded only 15 times.

In addition, we utilized a second data set \cite{TransferDataset} to apply transfer learning for the refined model. With a focus on the challenges we laid out, this second data set consists of flying objects at a noticeably farther distance than our initial data set. It consists of 11,998 images, where the average image size is 0.33 mp with a median image ratio of 640x512. The images are separated into a 90\% train and 10\% validation split. An image may contain more than one object and class, however, it has an average of one object per image, reaching a total count of 12,410 annotated objects. With only four different objects, each class is well represented: drones take up 38.8\% of the annotated objects, 21.2\% helicopters, 20.4\% airplanes, and 19.6\% birds. Although Roboflow reports a bird class, the images that contain birds are not labeled and are not included as a class in the transfer model. This dataset was published on Roboflow in 2022 by Ahmed Mohsen ~\cite{TransferDataset}, having only 5 downloads by the time of this paper.

We chose the YOLOv8 architecture under the assumption that it would provide us with the highest probability of success given the task. YOLOv8 is assumed to be the new state-of-the-art due to its higher mean average precisions (mAPs) and lower inference speed on the COCO dataset. However, an official paper has 
yet to be released. It also specifically performs better at detecting aerial objects [Figure \ref{fig:YOLOv8_average_mAP_against_cats}]. We implement the code for YOLOv8 from the Ultralytics repository. We decide to implement transfer learning and initialize our models with pre-trained weights to then begin training on the custom data set. These weights are from a model trained on the COCO dataset. Due to only having access to a single NVIDIA RTX 3080 and 3070, 
a greedy model selection/hyper-parameter tuning approach was chosen. We first train a version of the small, medium, and large versions of the model with default hyper-parameters for 100 epochs. Then, we decide which model is optimal for our use case given the trade-off between inference 
speed and mAP-50-95 on the validation set. After the model size is selected, a greedy hyper-parameter search is conducted with 10 epochs per each 
set of hyper-parameters. The model with the optimal hyper-parameters trains for 163 epochs to generate the generalized model. After this model learns abstract feature representations for a wide array of flying objects, we then transfer learn these weights to a data set that is more representative of the real world \cite{TransferDataset} to generate the refined model. This data set contains 3 classes: helicopter, plane, and drone, with very high variance in object spatial sizes. For evaluation, we are particularly interested in evaluating mAP50-95 and inference speed, as these are the most common measures of success across most object detection algorithms. Due to the large class imbalance, poor performance on the validation set was anticipated in the minority classes. However, this was not observed [Figure \ref{fig:confusion_matrix}].

Mean average precision (mAP) is one of the most used evaluation metrics for object detection. mAP takes the average precision (AP) over all classes and computes them at a pre-specified Intersection over Union (IoU) threshold. To define precision, we need to define true positives and false positives for object detection. A true positive will be determined when the IoU between the predicted box and ground truth is greater than the set IoU threshold, while a false positive will have the IoU below that threshold. Then, precision can be defined as $\dfrac{tp}{tp+fp}$. We take the mean over a class by iterating over a set of thresholds and averaging them. For mAP50-95, we take steps of 0.05 starting from an IoU threshold of 0.5 and stopping at 0.95. The average precision over this interval is the class AP. Do this for all classes and take the average over them and we generate the mAP50-95.

\subsection{Generalized Model Choice and Performance}

We evaluate small, medium, and large versions of the models to determine an optimal trade-off between inference speed and mAP50-95 to then optimize the hyper-parameters. The small, medium, and large models have (11151080, 25879480, \& 43660680) parameters and (225, 295, \& 365) layers respectively. After training the models, we see there is a noticeable increase in mAP50-95 between small and medium models (0.05), but not much delta between medium and large (0.002). We also see that small, medium, and large infer at 4.1, 5.7, and 9.3 milliseconds respectively on the validation set. However, our original goal is to reach an average inference speed between 30 to 60 frames for 1080p. When testing the medium-size model on multiple 1080p HD videos, we observe an average total speed (pre-process speed (0.5ms) + inference speed (17.25ms) + post-process speed (2ms)) of 19.75 ms (50 frames per second), which aligns with our primary objective. This leads to our selection of the medium-size model to begin tuning hyper-parameters.

\begin{figure}[!tbp]
  \centering

  \begin{minipage}[b]{0.4\textwidth}
    \hspace*{-1.5cm} 
    \includegraphics[width=9.5cm,height=7cm]{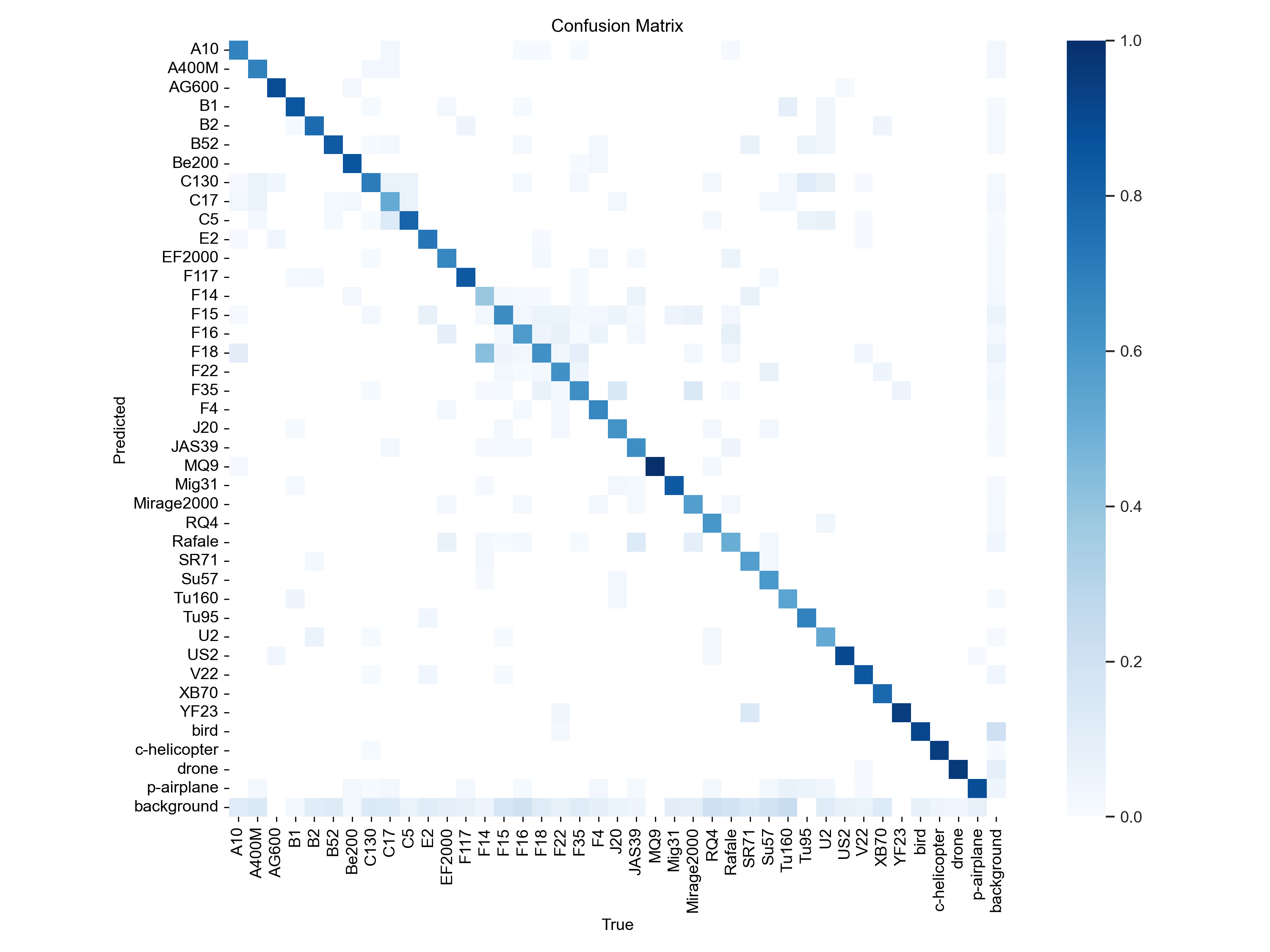}
    \caption{Confusion matrix for all classes.}
    \label{fig:confusion_matrix}
  \end{minipage}
  \hfill

  \begin{minipage}[b]{0.4\textwidth}
    \includegraphics[width=7cm,height=7.9cm]{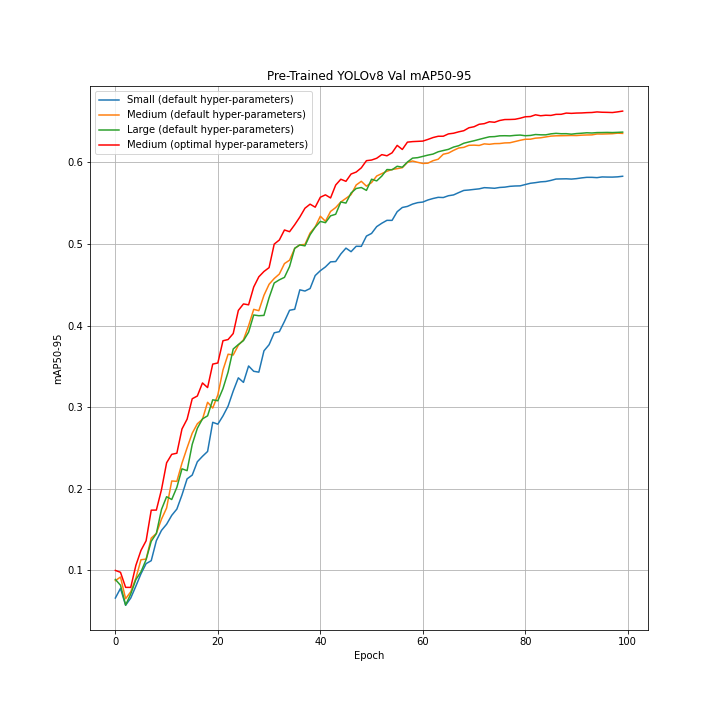}
    \caption{YOLOv8 validation mAP50-95.}
    \label{fig:YOLOv8_mAP50-95_val}
  \end{minipage}
  \label{fig:Model_Evaluation}
\end{figure}

Due to a lack of computational resources, we evaluate 10 epochs for each set of hyper-parameters as an indicator for the potential performance of additional epochs. We observe that this assumption is correct, as training with the optimal set of hyper-parameters achieves better performance at epoch 100 compared to default hyper-parameters (0.027) [Figure \ref{fig:YOLOv8_mAP50-95_val}]. We choose the best hyper-parameters based on validation mAP50-95 as batch size of 16, stochastic gradient descent (SGD) as the optimizer, momentum of 0.937, weight decay of 0.01, classification loss weight $\lambda_{cls}$ = 1, box loss weight $\lambda_{box}$ = 5.5, and distribution focal loss weight $\lambda_{dfl}$ = 2.5. After training for 163 epochs, we achieve a mAP50-95 of 0.685 and an average inference speed on 1080p videos of 50 fps.

\subsection{Loss Function and Update Rule}
The generalized loss function and weight update procedure can be defined as follows: 
\begin{equation}\label{Generalized Loss}
\mathcal{L}(\theta) = \dfrac{\lambda_{box}}{N_{pos}}\mathcal{L}_{box}(\theta) + \dfrac{\lambda_{cls}}{N_{pos}}\mathcal{L}_{cls}(\theta) + \dfrac{\lambda_{dfl}}{N_{pos}}\mathcal{L}_{dfl}(\theta) + \phi\Vert \theta \Vert_2^2
\end{equation}    
\begin{equation}\label{Velocity}
V^t = \beta V^{t-1} + \nabla_{\theta}\mathcal{L}(\theta^{t-1})
\end{equation}    
\begin{equation}\label{Weight Update}
\theta^{t} = \theta^{t-1} - \eta V^{t}
\end{equation}

(\ref{Generalized Loss}) is the generalized loss function incorporating the individual loss weights and a regularization term with weight decay $\phi$, (\ref{Velocity}) is the velocity term with momentum $\beta$, and (\ref{Weight Update}) is the weight update rule with $\eta$ as the learning rate. The specific YOLOv8 loss function can be defined as:
\begin{multline}
\mathcal{L} = \dfrac{\lambda_{box}}{N_{pos}}\sum_{x,y}\mathbbm{1}_{c^*_{x,y}}\big[1 - q_{x,y} + \dfrac{\Vert b_{x,y} - \hat{b}_{x,y} \Vert_2^2}{\rho^2} + \alpha_{x,y}\nu_{x,y}\big] \\  + \dfrac{\lambda_{cls}}{N_{pos}}\sum_{x,y}\sum_{c\in classes}y_clog(\hat{y}_c) + (1 - y_c)log(1 - \hat{y}_c) \\
+ \dfrac{\lambda_{dfl}}{N_{pos}}\sum_{x,y}\mathbbm{1}_{c^*_{x,y}}\Big[-(q_{(x,y)+1} - q_{x,y})log(\hat{q}_{x,y}) \\
+ (q_{x,y} - q_{(x,y)-1})log(\hat{q}_{(x,y)+1})\Big]
\end{multline}

where:
\begin{equation*}\label{IoU}
q_{x,y} = IoU_{x,y} = \dfrac{\hat{\beta}_{x,y}\displaystyle \cap\beta_{x,y}}{\hat{\beta}_{x,y}\displaystyle \cup\beta_{x,y}}
\end{equation*}
\begin{equation*}\label{v}
\nu_{x,y} = \dfrac{4}{\pi^2}(arctan(\dfrac{w_{x,y}}{h_{x,y}}) - arctan(\dfrac{\hat{w}_{x,y}}{\hat{h}_{x,y}}))^2
\end{equation*}
\begin{equation*}\label{a}
\alpha_{x,y} = \dfrac{\nu}{1 - q_{x,y}}
\end{equation*}
\begin{equation*}\label{y_hat}
\hat{y}_c = \sigma({\cdot})
\end{equation*}
\begin{equation*}\label{q_hat}
\hat{q}_{x,y} = softmax({\cdot})
\end{equation*}

and:
\begin{itemize}
\item $N_{pos}$ is the total number of cells containing an object.
\item $\mathbbm{1}_{c^*_{x,y}}$ is an indicator function for the cells containing an object. 
\item $\beta_{x,y}$ is a tuple that represents the ground truth bounding box consisting of ($x_{coord}$,$y_{coord}$, width, height).
\item $\hat{\beta_{x,y}}$ is the respective cell's predicted box.
\item $b_{x,y}$ is a tuple that represents the central point of the ground truth bounding box.
\item $y_c$ is the ground truth label for class c (not grid cell c) for each individual grid cell (x,y) in the input, regardless if an object is present.
\item $q_{(x,y)+/- 1}$ are the nearest predicted boxes IoUs (left and right) $\in c^*_{x,y}$.
\item $w_{x,y}$ and $h_{x,y}$ are the respective boxes width and height.
\item $\rho$ is the diagonal length of the smallest enclosing box covering the predicted and ground truth boxes.
\end{itemize}

Each cell then determines its best candidate for predicting the bounding box of the object. This loss function includes the complete IoU (CIoU) loss proposed by Zheng et al. \cite{CIoU} as the box loss, the standard binary cross entropy for multi-label classification as the classification loss (allowing each cell to predict more than 1 class), and the distribution focal loss proposed by Li et al. \cite{GFL} as the 3rd term.

\subsection{Model Confusion and Diagnosis}

\begin{figure*}[h]
    \centering
    \includegraphics[width=1.0\textwidth]{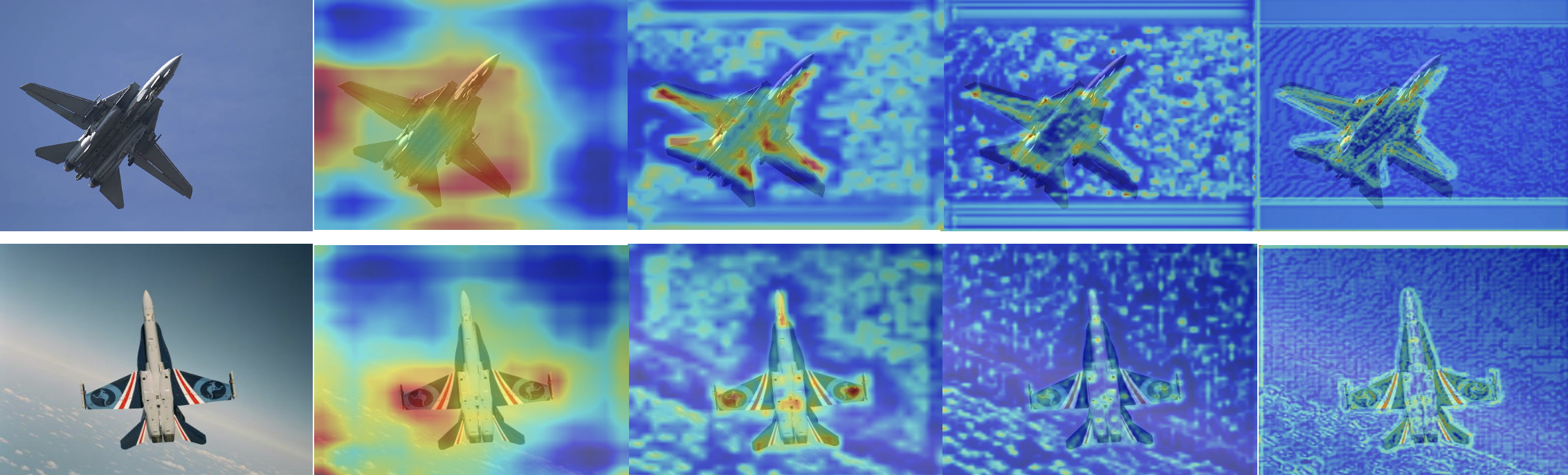}
    \caption{Feature activation maps for the F-14 and F-18 fighter jets. From left to right, we have the four stages of the model’s CSPDarkNet53 backbone.}
    \label{fig:F14vsF18}
\end{figure*}
\begin{figure*}[h]
    \centering
    \includegraphics[width=1.0\textwidth]{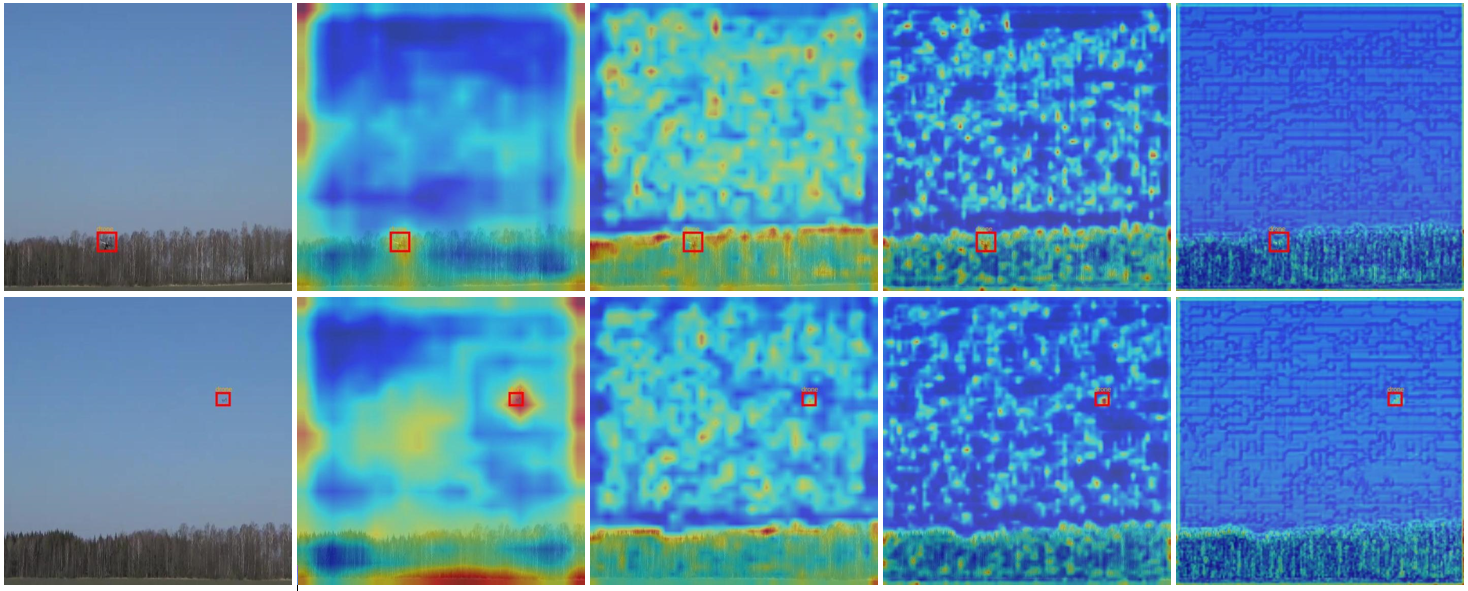}
    \caption{From left to right, (1) picks up the drone, (2) picks up tree top granularity - tree tops are more granular than stumps, (3) granular version of layer (2), (4) an outlier, texturized analysis of what the object is.}
    \label{fig:drone_gradcam}
\end{figure*}

One of the primary challenges in object detection is dealing with data sets with low inter-class variance, i.e. multiple classes that look similar to each other compared to the rest of the labels. Take, for example, the F-14 and F-18, which are displayed in Figure \ref{fig:F14vsF18}. Both have similar-looking wing shapes, two rudders, an engine, a cockpit, and a respective payload. In this confusion matrix [Figure \ref{fig:confusion_matrix}], the model is most likely to misclassify an F-14 as an F-18. This type of misclassification typically affects classes in categories with low inter-class variance amongst themselves. Visualizing activation maps \cite{MMYOLOViz} is a technique that helps us understand what pixels in the input image are important for determining its class. 

Generally, deeper layers in CNNs extract more granular/complex/low-level feature representations. YOLOv8 incorporates this idea into its architecture by having repeating modules and multiple detection heads when making its prediction. For our experimentation, we use MMYolo \cite{MMYOLOViz} to create activation maps at different stages of our backbone. We expect some sense of differentiation in the different feature maps. If our model shows similar feature activations for F-14s and F-18s, we can say that may be the reason for class confusion.

MMYolo \cite{MMYOLOViz} by Yamaguchi et al. is an open-source toolbox for YOLO series algorithms based on PYTorch. MMYolo can decompose the most popular YOLO algorithms, making them easily customizable and ready for analysis. For our analysis, we employed MMYolo to first convert the weights from .pt (Pytorch model) to .pth (State dictionary file, i.e., weights, bias, etc.), and to second visualize the different activation maps of YOLOv8 during inference. MMYolo allows you to specify the model type, weight file, target layer, and channel reduction.

YOLOv8 uses CSPDarknet53 \cite{darkNet} as its backbone [Figure \ref{fig:YOLOv8_arch}], a deep neural network that extracts features at multiple resolutions (scales) by progressively down-sampling the input image. The feature maps produced at different resolutions contain information about objects at different scales in the image and different levels of detail and abstraction. YOLOv8 can incorporate different feature maps at different scales to learn about object shapes and textures, which helps it achieve high accuracy in most object detection tasks. YOLOv8's backbone consists of four sections, each with a single convolution followed by a c2f module \cite{YOLOv8Website}. The c2f module is a new introduction to CSPDarknet53. The module comprises splits where one end goes through a bottleneck module (two 3x3 convolutions with residual connections). The bottleneck module output is further split N times, where N corresponds to the YOLOv8 model size. These splits are all finally concatenated and passed through one final convolution layer. This final layer is where we will get the activations.

\begin{figure*}[h]
    \centering
    \subfloat[\centering Generalized Model \label{fig:GeneralizedModel}]{{ \includegraphics[width=1 \textwidth]{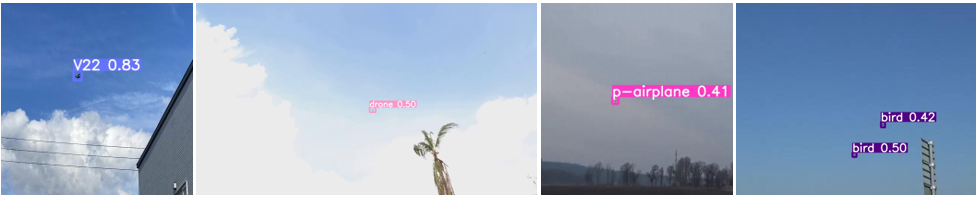}}}\hspace{0.1em}
    \centering    
    \subfloat[\centering Refined Model - Transfer Learning Images \label{fig:RefinedModel}]{{\includegraphics[width=1 \textwidth]{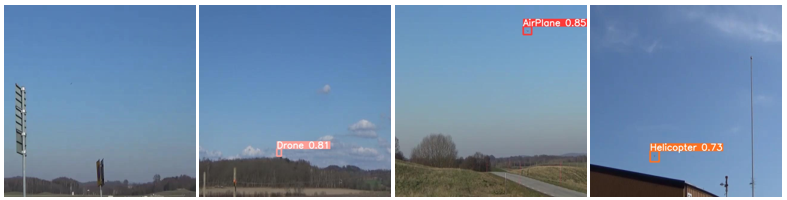}}}    
    \caption{Prediction Images}%
    \label{fig:PredictionModel}
\end{figure*}

Figure \ref{fig:F14vsF18} shows the original F-14 and F-18 images and the activations of the four c2f stages in the network, with each stage being more profound in the network from the second image right. The activation map corresponding to the shallowest c2f module shows the broadest activation. This module detects the two wings of the aircraft and determines that this object is a plane. The second activation map corresponds to the second c2f module in our backbone. It shows strong activations at different components of the aircraft, such as locating the wings, body, cockpit, and payload. It appears that this layer is attempting to infer what kind of aircraft is being presented in the image by highlighting these features. The third activation map is starting to dive into the individual textures of the components of the aircraft, presumably checking for minute differences in the jet's structure. Finally, the model's final c2f module activates extremely fine-grained details and outlines in the respective images.  These similar feature activation maps could be the reason that the model confuses the two.

\section{Results}

\subsection{Generalized Model}
To highlight our results, we address three challenging conditions: (1) detecting and classifying extremely small objects, (2) identifying flying objects that blend into their background, and (3) classifying different types of flying objects. We examined the performance of our generalized model, \cite{InitialDataset}, against these challenges. This is demonstrated in Figure \ref{fig:PredictionModel}, which features four images that represent the bird, drone, passenger airplane, and V22 classes.

The first of the four images showcases the model's ability to identify distant birds. In the second image, the model was put to the test against a very small drone that occupied only 0.026\% of the image size while also blending in with its background. The model still resulted in the correct detection and classification of the drone. The third image shows the model's ability to identify a minute passenger airplane of size 0.063\% of the image, which is also blended into its surroundings. Finally, the fourth image features a V22 aircraft, which is an underrepresented class and accounts for only 3.57\% of the entire dataset. A V22 can easily be mistaken as a drone due to its vertical propeller positioning. Despite these two hindering characteristics and only taking up 0.14\% of the entire image, the image exhibits the model's ability to still identify the V22 with impressive accuracy, achieving a confidence score of 0.83.

Despite the visual similarities between the birds, drones, and passenger airplanes in these images, our model successfully classified them with adequate confidence. These results illustrate our model's ability to overcome our identified challenges associated with object detection in real-world conditions, and also demonstrate our success in creating a solution that effectively tackles these challenges. Overall, it does very well at distinguishing various types of flying objects despite the need to account for multiple different classes of aircraft.

\subsection{Refined Model}
To generate the refined model, we initialized the model with the weights learned from the generalized model and default hyperparameters. We then trained the model on the  ``real world`` data set for 199 epochs \cite{TransferDataset}. This data set was selected to focus on our challenge of detecting and classifying extremely small objects in appearance. Figure \ref{fig:PredictionModel} displays our results, featuring four distinct images that represent the bird, drone, airplane, and helicopter objects.

The first image contains an extremely small bird that only takes up 0.02\% of the image. Even with the lack of the bird class in our training process, our model correctly identified that the object was not any of the other available classes, even while allowing a very low confidence threshold of 0.20. The second image contains a drone, which also only took up 0.02\% of its image. This drone is nearly indistinguishable from the background clouds to the human eye, yet our model was still able to classify it with a confidence score of 0.81. The third image includes a small airplane that takes up 0.034\% of pixels, which our model was still able to correctly identify and classify with a high confidence score of 0.85. In the final image, a barely visible helicopter (0.01\% of the image) was correctly classified with a confidence score of 0.73.

In Figure \ref{fig:drone_gradcam}, we can see that the feature map activation correctly segments the object in the first layer. The second layer starts picking out all of the tree tops, which can be explained by the higher relative variance of the tree tops. In the third layer, we see more importance being placed on the background and more granular features being detected. In the fourth layer, we see the outline of the drone itself. In the second row, the true strength of the localization accuracy with an over-emphasized detection is displayed. In the second layer, we see a de-emphasis on the background. In the third and fourth layers, we see the same behavior as before.

Our model achieves state-of-the-art results, achieving a mAP50 of 0.991 and mAP50-95 of 0.835 across the plane, helicopter, and drone classes. These results demonstrate that our generalized model serves as an excellent base for transfer learning, particularly when dealing with extremely small objects, blended backgrounds, and distinguishing between drones and other flying objects. 

\section{Discussion}

To the problem of flying object detection, we apply transfer learning with weights learned from our generalized model to our refined model in order to achieve state-of-the-art results in this domain. We argue that our algorithm extracts better feature representations of flying objects than those seen in previous research, furthering the current state of research in this domain. Our refined model achieves a 99.1\% mAP50, 98.7\% Precision, and 98.8\% Recall with 50 fps inference speed on the 3-class data set (drone, plane, and helicopter), surpassing models generated from previous research to a significant extent. Aydin et al. proposed a YOLOv5 instance that achieved 90.40\% mAP50, 91.8\% Precision, and 87.5\% Recall with 31 fps inference speed trained on a data set only containing drones and birds \cite{Yolov5Drone}. Rozantsev et al. trained their proposed model on a data set reflective of ours, containing flying objects that occupy small portions of the input image with clustered backgrounds. They achieve an 84.9\% AP on a data set containing only UAVs and 86.5\% AP on a data set containing only aircraft \cite{SingleMovingDetector}. Al-Qubaydhi et al. proposed a model utilizing the YOLOv5 framework and achieves an impressive 94.1\% mAP50, 94.7\% Precision, and 92.5\% Recall on a dataset containing only one class of drones. \cite{UAVYOLOv5Transfer}. Even with our exceptional results, a potential limitation of our refined model is that it was trained on a data set with a low amount of distinct environments. To address this potential generalization issue, we suggest utilizing our generalized model weights to transfer learn on a data set with higher frequency of distinct backgrounds.

\section{Model Architecture}
With the publication of “You Only Look Once: Unified, Real-Time Object Detection” first proposed by Redmon et al. \cite{YOLO_OG} in 2015, one of the most popular object detection algorithms, YOLOv1, was first described as having a “refreshingly simple” approach \cite{CompReview}. At its inception, YOLOv1 could process images at 45 fps, while a variant, fast YOLO, could reach upwards of 155 fps. It also achieved high mAP compared to other object detection algorithms at the time.\\
\indent The main proposal from YOLO is to frame object detection as a one-pass regression problem. YOLOv1 comprises a single neural network, predicting bounding boxes and associated class probability in a single evaluation. The base model of YOLO works by first dividing the input image into an S x S grid where each grid cell (i,j) predicts B bounding boxes, a confidence score for each box, and C class probabilities. The final output will be a tensor of shape S x S x (B x 5 + C).

\subsection{YOLOv1 Overview}
\begin{figure}[h]
    \centering
    \includegraphics[width=0.45\textwidth]{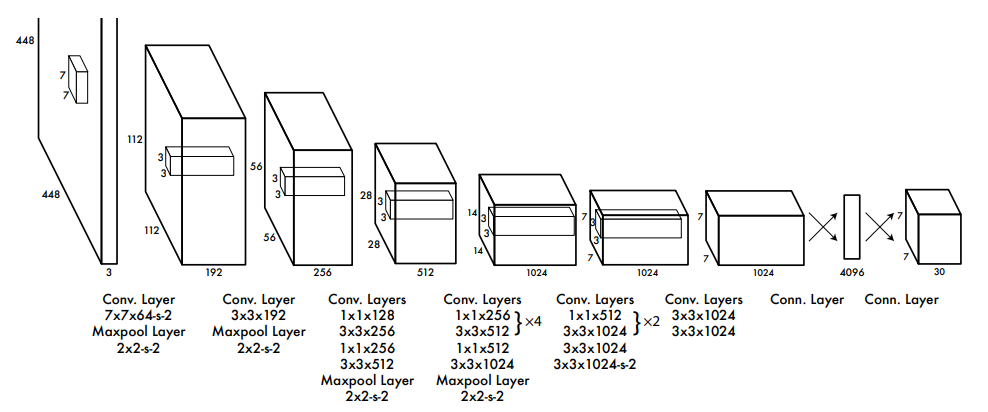}
    \caption{YOLO Architecture \cite{YOLO_OG}}
    \label{fig:YOLOv1Architecture}
\end{figure}

YOLOv1 architecture [Figure \ref{fig:YOLOv1Architecture}] consists of 24 convolutional layers followed by two fully connected layers. In the paper \cite{YOLO_OG}, the authors took the first 20 convolutional layers from the backbone of the network and, with the addition of an average pooling layer and a single fully connected layer, it was pre-trained and validated on the ImageNet 2012 dataset. During inference, the final four layers and 2 FC layers are added to the network; all initialized randomly.

YOLOv1 uses stochastic gradient descent as its optimizer. The loss function, shown by Equation \ref{YOLO_Loss_equation}, comprises two parts: localization loss and classification loss. The localization loss measures the error between the predicted bounding box coordinates and the ground-truth bounding box. The classification loss measures the error between the predicted class probabilities and the ground truth. The $\lambda_{coord}$ and $\lambda_{noobj}$ are regularization coefficients that regulate the magnitude of the different components, emphasizing object localization and de-emphasizing grid cells without objects.

\begin{multline} \label{YOLO_Loss_equation}
\lambda_{coord}\sum_{i=0}^{S^2}\sum_{j=0}^{B}\mathbbm{1}_{ij}^{obj}\Big[(x_i-\hat{x}_i)^2+(y_i-\hat{y}_i)^2\Big] \\ + \lambda_{coord}\sum_{i=0}^{S^2}\sum_{j=0}^{B}\mathbbm{1}_{ij}^{obj}\Big[(\sqrt{w_i}-\sqrt{\hat{w}_i})^2+(\sqrt{h_i}-\sqrt{\hat{h}_i})^2\Big]\\
+\sum_{i=0}^{S^2}\sum_{j=0}^{B}\mathbbm{1}_{ij}^{obj}(C_i-\hat{C}i)^2\\
+\lambda_{noobj}\sum_{i=0}^{S^2}\sum_{j=0}^{B}\mathbbm{1}_{ij}^{noobj}(C_i-\hat{C}i)^2\\
+\sum_{i=0}^{S^2}\mathbbm{1}_{i}^{obj}\sum_{c\in classes}(p_i(c)-\hat{p}_i(c))^2
\end{multline}

\begin{figure*}[h]
    \centering
    \includegraphics[width=1.0\textwidth]{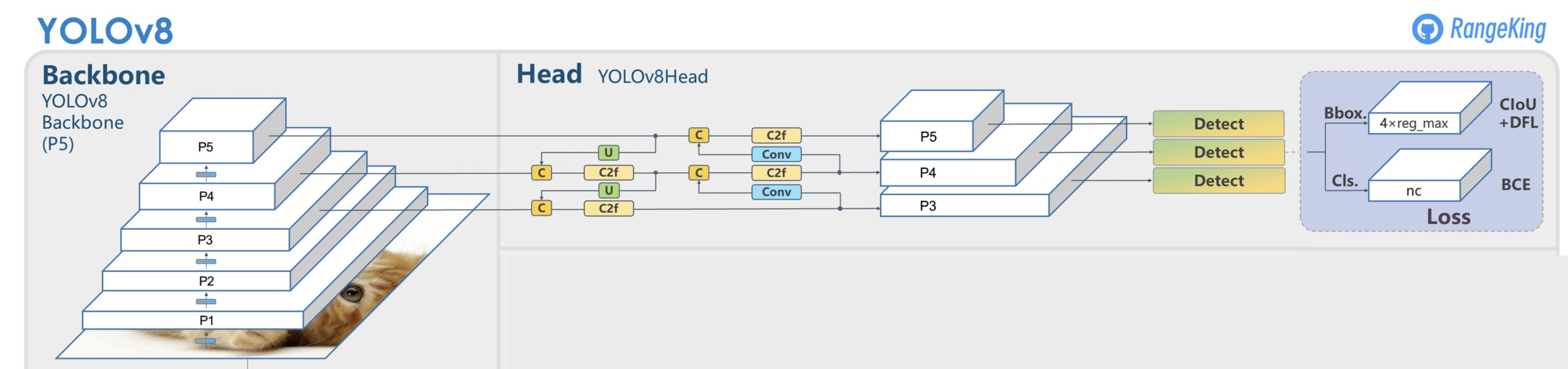}
    \caption{YOLOv8 Architecture ~\cite{YOLOv8Website}}
    \label{fig:YOLOv8_arch}
\end{figure*}

\subsection{YOLOv5 Overview}

YOLOv5 \cite{Drone-Detection-Using-YOLOv5} is an object detection model introduced in 2020 by Ultralytics, the originators of the original YOLOv1 and YOLOv3. YOLOv5 achieves state-of-the-art performance on the COCO benchmark dataset \cite{YOLOv5Doc} while also being fast and efficient to train and deploy. YOLOv5 made several architectural changes, most notably the standardized practice of structuring the model into three components: the backbone, neck, and head.

The backbone of YOLOv5 is Darknet53, a new network architecture that focuses on feature extraction characterized by small filter windows and residual connections. Cross-stage partial connections (CSP) enable the architecture to achieve a richer gradient flow while reducing computation as proposed by Wang et al. \cite{cspNET}. 

The neck \cite{CompReview}, as described by Teven et al., of YOLOv5 connects the backbone to the head, whose purpose is to aggregate and refine the features extracted by the backbone, focusing on enhancing the spatial and semantic information across different scales. A Spatial Pyramid Pooling (SPP) \cite{SPP} module removes the fixed-size constraint of the network, which removes the need to warp, augment, or crop images. This is followed by a CSP-Path Aggregation Network \cite{cspNET} module, which incorporates the features learned in the backbone and shortens the information path between lower and higher layers.

YOLOv5’s head consists of three branches, each predicting a different feature scale. In the original publication of the model \cite{YOLOv5Doc}, the creators used three grid cell sizes of 13 x 13, 26 x 26, and 52 x 52, with each grid cell predicting B = 3 bounding boxes. Each head produces bounding boxes, class probabilities, and confidence scores. Finally, the network uses Non-maximum Suppression (NMS) \cite{NMS} to filter out overlapping bounding boxes.

YOLOv5 incorporates anchor boxes, which are fixed-sized bounding boxes used to predict the location and size of objects within an image. Instead of predicting arbitrary bounding boxes for each object instance, the model predicts the coordinates of the anchor boxes with predefined aspect ratios and scales and adjusts them to fit the object instance.

\subsection{YOLOv8 Overview}
YOLOv8 is the latest version of the YOLO object detection model. This latest version has the same architecture as its predecessors [Figure \ref{fig:YOLOv8_arch}], but it introduces numerous improvements compared to the earlier versions of YOLO, such as a new neural network architecture that utilizes both Feature Pyramid Network (FPN) and Path Aggregation Network (PAN) and a new labeling tool that simplifies the annotation process. This labeling tool contains several useful features like auto labeling, labeling shortcuts, and customizable hotkeys. The combination of these features makes it easier to annotate images for training the model.

The FPN works by gradually reducing the spatial resolution of the input image while increasing the number of feature channels. This results in the creation of feature maps that are capable of detecting objects at different scales and resolutions. The PAN architecture, on the other hand, aggregates features from different levels of the network through skip connections. By doing so, the network can better capture features at multiple scales and resolutions, which is crucial for accurately detecting objects of different sizes and shapes ~\cite{CompReview}.

\subsection{YOLOv8 vs YOLOv5}
\begin{figure}[!tbp]
  \centering
  \begin{minipage}[b]{0.4\textwidth}
    \includegraphics[width=\textwidth]{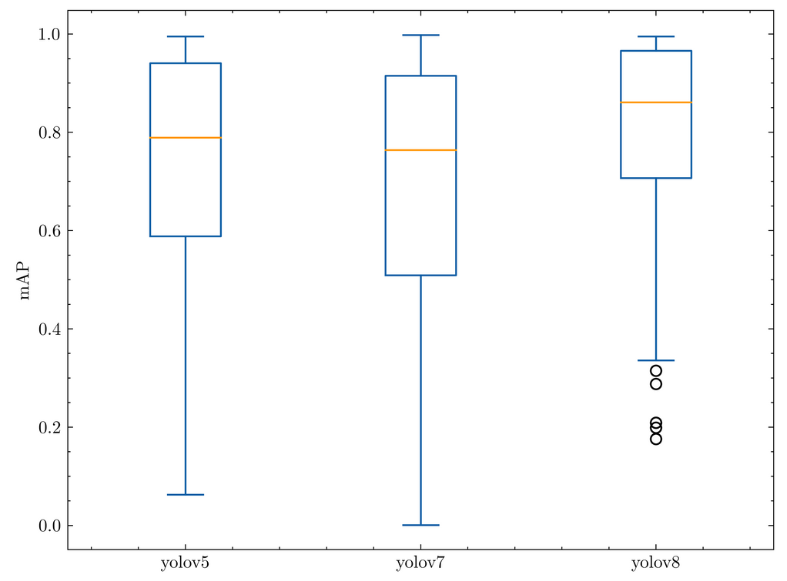}
    \caption{YOLOs mAP@.50 against RF100.}
    \label{fig:YOLOv8_mAP}
  \end{minipage}
  \hfill
  \begin{minipage}[b]{0.4\textwidth}
    \includegraphics[width=\textwidth]{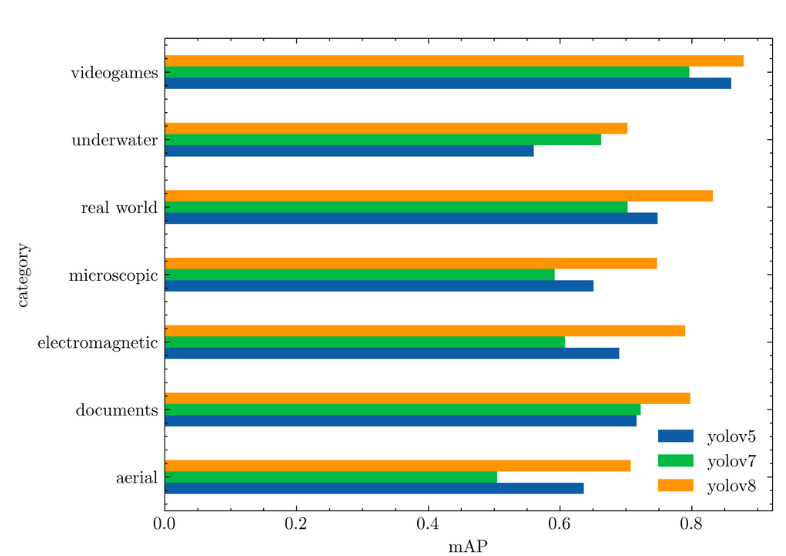}
    \caption{YOLOs average mAP@.50 against RF100 categories}
    \label{fig:YOLOv8_average_mAP_against_cats}
  \end{minipage}
  \label{fig:Model_Evaluation2}
\end{figure}

The reason why YOLOv8 is being compared to YOLOv5 rather than any other version of YOLO is that YOLOv5’s performance and metrics are closer to YOLOv8’s. However, YOLOv8 surpasses YOLOv5 in aspects including a better mAP as seen in Figure \ref{fig:YOLOv8_average_mAP_against_cats}. Along with a better mAP, this shows that YOLOv8 has fewer outliers when measured against the RF100. RF100 is a 100-sample dataset from the Roboflow universe, which is a repository of 100,000 data sets. We also witness YOLOv8 outperforming YOLOv5 for each RF100 category. From Figure \ref{fig:YOLOv8_mAP}, we can see that YOLOv8 produces similar or even better results compared to YOLOv5 ~\cite{YOLOv8Website}.

As mentioned previously, YOLOv8 uses a new architecture that combines both FAN and PAN modules. FPN is used to generate feature maps at multiple scales and resolutions, while PAN is used to aggregate features from different levels of the network to improve accuracy. The results of the combined FAN and PAN modules are better than YOLOv5, which uses a modified version of the CSPDarknet architecture. This modified version of CSPDarknet is based on cross-stage partial connections, which improves the flow of information between different parts of the network.

Another difference the two models have is based on their training data. YOLOv8 was trained on a larger and more diverse dataset compared to YOLOv5. YOLOv8 was trained on a blend of the COCO dataset and several other datasets, while YOLOv5 was trained primarily on the COCO dataset. Because of this, YOLOv8 has a better performance on a wider range of images.

YOLOv8 includes a new labeling tool called RoboFlow Annotate, which is used for image annotation and object detection tasks in computer vision. RoboFlow Annotate makes it easier to annotate images for training the model and includes several features such as auto labeling, labeling shortcuts, and customizable hotkeys. In contrast, YOLOv5 uses a different labeling tool called LabelImg. LabelImg is an open-source graphical image annotation tool that allows its users to draw bounding boxes around objects of interest in an image, and then export the annotations in the YOLO format for training the model.

YOLOv8 includes more advanced post-processing techniques than YOLOv5, which is a set of algorithms applied to the predicted bounding boxes and objectiveness scores generated by the neural network. These techniques help to refine the detection results, remove redundant detections, and improve the overall accuracy of the predictions. YOLOv8 uses Soft-NMS, a variant of the NMS technique used in YOLOv5. Soft-NMS applies a soft threshold to the overlapping bounding boxes instead of discarding them outright, whereas NMS removes the overlapping bounding boxes and keeps only the ones with the highest objectiveness score.

Output heads refer to the final layers of a neural network that predict the locations and classes of objects in an image. In YOLO architecture, there are normally several output heads that are responsible for predicting different aspects of the detected objects, such as the bounding box coordinates, class probabilities, and objectiveness scores. These output heads are typically connected to the last few layers of the neural network and are trained to output a set of values that can be used to localize and classify objects in an image. The number and type of output heads used can vary depending on the specific object detection algorithm and the requirements of the task at hand. YOLOv5 has three output heads while YOLOv8 has one output head. YOLOv8 does not have small, medium, and large anchor boxes. It uses an anchor-free detection mechanism that directly predicts the center of an object instead of the offset from a known anchor box. This reduces the number of box predictions and speeds up the post-processing process in return.

It is fair to note that YOLOv8 is slightly slower than YOLOv5 in regard to object detection speed. However, YOLOv8 is still able to process images in real-time on modern GPUs.

Both YOLOv5 and YOLOv8 use mosaic augmentation on the training set. Mosaic augmentation is a data augmentation technique that takes four random images from the training set and combines them into a single mosaic image. This image, where each quadrant contains a random crop from one of the four input images, is then used as input for the model ~\cite{MosaicAug}.


\clearpage
\newpage
														 
{\small
\bibliographystyle{ieee_fullname}
\bibliography{egbib}
}
\end{document}